\pdfoutput=1

\documentclass[11pt]{article}
\usepackage[toc,page]{appendix}
\usepackage[preprint]{acl}

\usepackage{times}
\usepackage{latexsym}
\usepackage{xcolor}
\usepackage[T1]{fontenc}

\usepackage{amsmath}

\usepackage{pgfplots}
\usepackage{subcaption}
\pgfplotsset{compat=newest}
\usepackage{tikz}

\usepackage[utf8]{inputenc}

\usepackage{microtype}
\usepackage{mathrsfs}
\usepackage{amsmath}
\usepackage{bm}
\usepackage{multirow}
\usepackage{adjustbox}
\usepackage{graphicx}
\usepackage{colortbl}
\usepackage{arydshln}
\usepackage{microtype} 
\usepackage{amsmath, amssymb, mathtools}
\usepackage{appendix}
\usepackage{wrapfig}
\usepackage{xcolor}
\usepackage{tabularray}
\usepackage{inconsolata}

\usepackage{hyperref}       
\usepackage{url}            
\usepackage{booktabs}       
\usepackage{amsfonts}       
\usepackage{nicefrac}       
\usepackage{microtype}      
\usepackage{xcolor}         
\usepackage{tcolorbox}
\usepackage{amsmath}
\usepackage{amsthm}
\usepackage{graphicx}
\usepackage{mathtools}
\usepackage{cleveref}
\usepackage{algorithm}
\usepackage[noend]{algpseudocode} 
\usepackage{wrapfig}

\usepackage{graphicx}

\title{Does a Global Perspective Help Prune Sparse MoEs Elegantly?}

\author{Zeliang Zhang$^{1}$\,  Nikhil Ghosh$^{2}$\,  Jiani Liu$^{1}$\, Bin Yu$^{3}$\,  Xiaodong Liu$^{4}$\ \\
$^{1}$University of Rochester\quad
$^{2}$Flatiron Institute\quad 
$^{3}$University of California, Berkeley\quad \\
$^{4}$Microsoft Research\quad
 \\\footnotesize{\texttt{\{zeliang.zhang, jiani.liu\}@rochester.edu,}}
 \footnotesize{\texttt{nikhil\_ghosh@berkeley.edu}} \\ 
\footnotesize{\texttt{binyu@stat.berkeley.edu},}
\footnotesize{\texttt{xiaodl@microsoft.com}}}

\begin{document}
\maketitle
\begin{abstract}
Empirical scaling laws for language models have encouraged the development of ever-larger LLMs, despite their growing computational and memory costs. Sparse Mixture-of-Experts (MoEs) offer a promising alternative by activating only a subset of experts per forward pass, improving efficiency without sacrificing performance. However, the large number of expert parameters still leads to substantial memory consumption.

Existing pruning methods typically allocate budgets uniformly across layers, overlooking the heterogeneous redundancy that arises in sparse MoEs. We propose \textbf{GRAPE} (\textbf{G}lobal \textbf{R}edundancy-\textbf{A}ware \textbf{P}runing of \textbf{E}xperts), a global pruning strategy that dynamically allocates pruning budgets based on cross-layer redundancy. Experiments on Mixtral-8x7B, Mixtral-8x22B, DeepSeek-MoE, Qwen-MoE, and GPT-OSS show that, under the same pruning budget, GRAPE consistently achieves the best average performance. On the three main models reported in the paper, it improves average accuracy over the strongest local baseline by 1.40\% on average across pruning settings, with gains of up to 2.45\%.
\end{abstract}

\section{Introduction}
Supported by the scaling law~\citep{kaplan2020scaling}, increasing the number of parameters enhances the capacity of large language models (LLMs), leading to impressive yet sometimes spurious performance across various tasks~\citep{chang2024survey}. However, this growth also introduces significant computational overhead during both training and inference~\citep{li2024personal}. In recent years, sparse mixture-of-experts (MoEs)~\citep{zoph2022st,chen2023sparse} have emerged as an effective solution by replacing a single feed-forward network (FFN) with multiple expert layers. By sparsely activating different experts at each forward pass, MoEs reduce computation costs during inference while maintaining performance comparable to dense LLMs~\citep{pan2024dense}. Despite this advantage, MoEs introduce a noticeable memory cost~\citep{zhang2024diversifying}.


Many studies have explored effective strategies for pruning MoEs, which can be broadly categorized into four types: visiting frequency-guided~\citep{chen2022task,he2024demystifying}, router-guided~\citep{li24merge}, search-based~\citep{lu2024not}, and feature-based~\citep{zhang2024diversifying} methods. The core idea behind these approaches is to identify pairs of experts with similar behavior, allowing some to be safely removed or merged. However, these methods typically allocate the pruning budget uniformly across all layers, ignoring inter-layer variation in sparsity.

Motivated by the observation that redundancy varies substantially across MoE layers, we propose GRAPE (Global Redundancy-Aware Pruning of Experts), a global pruning method that dynamically allocates pruning budgets according to cross-layer redundancy. Rather than pruning the same number of experts per layer, our method adjusts the allocation to leverage layerwise sparsity differences, aiming to better balance memory reduction with preservation of model performance.

To validate the effectiveness of our approach, we apply it to Mixtral-8x7B/22B~\citep{jiang2024mixtral}, Deepseek-MoE~\citep{dai2024deepseekmoe}, Qwen-MoE~\citep{yang2024qwen2}, and GPT-oss~\citep{agarwal2025gpt},  under various global pruning budgets. Experimental results show that GRAPE consistently outperforms uniform layer-wise pruning baselines, highlighting the importance of accounting for cross-layer redundancy when pruning sparse MoEs.

\section{Background}

 There has been a growing body of work focused on pruning sparse MoEs. \citet{chen2022task} propose pruning less frequently visited experts based on task-specific usage. \citet{chowdhury2024provably} observe that less important experts tend to exhibit smaller changes in routing weights during fine-tuning. \citet{li24merge} suggest merging experts that are frequently visited by similar token groups in the fine-tuned dataset. \citet{he2024demystifying} explore pruning based on visitation frequency using a task-agnostic calibration dataset. \citet{lu2024not} identify redundant expert groups by analyzing the loss landscape on the calibration set. \citet{zhang2024diversifying} merge experts with similar output activations or weight parameters. \citet{lee2024stun} introduce a two-stage approach that first drops experts and then applies unstructured pruning for further efficiency. \citet{liu2024efficient} employ an evolutionary strategy to search for prunable expert subsets using a small task-specific calibration dataset. 


\section{Methodology}

\subsection{Preliminary}

Consider a large language model with $L$ sparse MoE layers. The output of the $l$-th MoE layer is given by
\begin{equation}\small
\begin{aligned}
y^l = \sum_{s_i \in \mathcal{S}} \alpha_{s_i}^l \cdot \phi_{s_i}^l(x),
\end{aligned}
\label{eq:moe_layer}
\end{equation}
where $x$ denotes the input representation, $\mathcal{S}=\{s_1,s_2,\ldots,s_N\}$ is the set of activated experts, $\phi_{s_i}^l$ denotes the $s_i$-th activated expert in layer $l$, and $\alpha_{s_i}^l$ is its corresponding routing coefficient. Each expert $\phi^l(\cdot)$ consists of two linear layers with a GeLU activation in between.

\subsection{Not all MoE layers are equally redundant}

Prior studies have highlighted the presence of expert redundancy within individual MoE layers. For example, \citet{zhang2024diversifying} use Central Kernel Alignment (CKA) to empirically assess intra-layer redundancy. Beyond this intra-layer phenomenon, we further observe that the degree of redundancy varies substantially across layers.

To formalize this observation, we define an expert similarity matrix for each MoE layer. Let $D^l \in \mathbb{R}^{N \times N}$ denote the pairwise similarity matrix of experts in the $l$-th MoE layer, where $D_{ij}^l$ measures the similarity between expert $i$ and expert $j$. In practice, $D^l$ can be instantiated using CKA~\citep{davarireliability}, mean squared error, or other similarity measures.

Based on $D^l$, we define the average intra-layer redundancy score as
\begin{equation}
\begin{aligned}
R^l = \frac{1}{N(N-1)} \sum_{i \neq j} D_{ij}^l,
\end{aligned}
\end{equation}
which captures the average pairwise similarity among experts in layer $l$.

To compare redundancy across layers, we further normalize the scores:
\begin{equation}
\begin{aligned}
\widetilde{R}^l = \frac{R^l - \min_{l'} R^{l'}}{\max_{l'} R^{l'} - \min_{l'} R^{l'}}.
\end{aligned}
\end{equation}
Here, $\widetilde{R}^l \in [0,1]$ represents the \emph{relative redundancy} of layer $l$, where a larger value indicates that experts in this layer are more redundant relative to those in other layers.

\begin{figure}
    \centering
    \includegraphics[width=\linewidth]{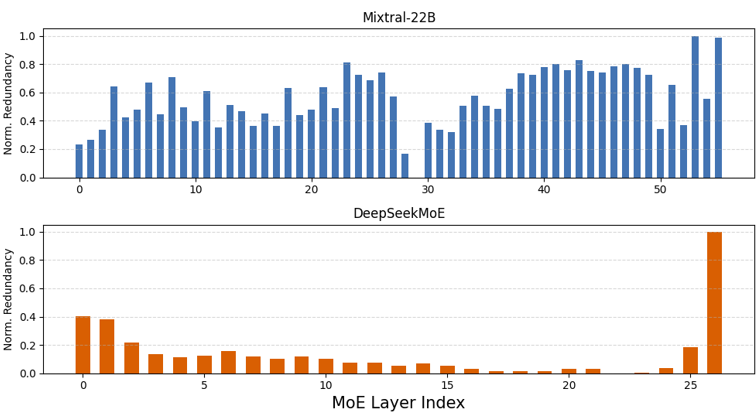}
    \caption{Cross-layer redundancy of different MoE models, including Mixtral-8x22B and Deepseek-MoE.}
    \label{fig:x_layer_redundancy}
\end{figure}

We visualize the cross-layer redundancy of different MoE models in \cref{fig:x_layer_redundancy}. As shown, expert redundancy varies substantially across layers within the same model. In general, earlier MoE layers tend to exhibit lower redundancy than later ones. However, this trend is not strictly monotonic, as some intermediate layers also display relatively low redundancy. These observations suggest that pruning strategies for MoEs should account for heterogeneous redundancy across layers, rather than applying a uniform pruning rule to all layers.

\subsection{Globally Pruning the MoEs}

To reduce expert redundancy across the model, we propose GRAPE, a global pruning strategy that explicitly accounts for cross-layer differences in redundancy. Unlike layer-wise pruning methods that remove a fixed number of experts from each layer independently, our approach jointly determines expert merging across all layers under a unified pruning budget.

Our objective is to reduce the total number of experts to a target value $K < LN$ by pruning structurally redundant experts. This corresponds to globally pruning exactly $LN-K$ experts. To formalize this, we construct a block-diagonal similarity matrix
\begin{equation}\small
\begin{aligned}
A = \mathrm{blockdiag}(D^1, D^2, \ldots, D^L) \in \mathbb{R}^{LN \times LN},
\end{aligned}
\end{equation}
where each block $D^l \in \mathbb{R}^{N \times N}$ encodes the pairwise similarity between experts in the $l$-th MoE layer.

We then consider the following objective:
\begin{equation}\small
\begin{aligned}
\arg\min_{M} \sum_{i \neq j} A_{ij}\cdot M_{ij},
\end{aligned}
\label{eq:global_optim_meta}
\end{equation}
where $M = \mathrm{blockdiag}(M^1, M^2, \ldots, M^L) \in \{0,1\}^{LN \times LN}$ is a masking matrix indicating the remaining experts in the pruned model.

However, directly optimizing \cref{eq:global_optim_meta} may lead to a degenerate solution. As illustrated in \cref{fig:x_layer_redundancy}, certain layers, such as the final layer of Deepseek-MoE, exhibit disproportionately high redundancy. In such cases, the pruning budget may be allocated excessively to only a few highly redundant layers, causing severe layer imbalance and even model collapse.

To mitigate this issue, we introduce a regularization term based on the \textbf{global entropy}, which characterizes how the retained experts are distributed across layers. Specifically, let
\begin{equation}
p_l = \frac{\mathcal{I}(D^l \odot M^l)}{\mathcal{I}(A \odot M)},
\end{equation}
where $\mathcal{I}(\cdot)$ is a counting function that returns the number of remaining experts in the corresponding layer or in the whole model. Then $p_l$ is the fraction of retained experts in layer $l$, with $\sum_l p_l = 1$.

Based on this layer-wise fraction, we define the global entropy as
\begin{equation}\small
\begin{aligned}
E = - \sum_l p_l \log p_l,
\end{aligned}
\label{eq:global_entropy}
\end{equation}
which is exactly the entropy of the distribution $\{p_l\}$. A larger entropy indicates that the retained experts are distributed more evenly across layers, whereas a smaller entropy indicates that pruning is overly concentrated in only a few layers.

In practice, rather than directly optimizing an entropy-regularized objective, GRAPE uses global entropy as a safeguard in a greedy pruning procedure. Specifically, we maintain the current clustering structure $\{\mathcal{C}^l\}_{l=1}^L$, where $\mathcal{C}^l$ denotes the current set of expert clusters in layer $l$, and each cluster corresponds to one retained expert after merging. Initially, each expert forms a singleton cluster. The global entropy is then computed from the layer-wise cluster fractions. Let
\begin{equation}
p_l = \frac{|\mathcal{C}^l|}{\sum_{l'} |\mathcal{C}^{l'}|},
\end{equation}
and define
\begin{equation}
E = - \sum_l p_l \log p_l.
\end{equation}

At initialization, we compute the entropy of the unpruned model and define an entropy threshold
\begin{equation}
\widehat{E} = E(1-\gamma),
\end{equation}
where $\gamma \in [0,1]$ is an entropy tolerance parameter. A larger $\gamma$ allows more imbalance across layers during pruning, while a smaller $\gamma$ enforces a more even allocation of retained experts.

We further maintain a layer-redundancy score
\begin{equation}
R^l = \sum_{i \neq j} D_{ij}^l,
\end{equation}
which quantifies the total residual similarity mass in layer $l$. Based on these quantities, GRAPE performs one-shot entropy-aware greedy pruning with restart, as summarized in \cref{alg:entropy_greedy}.

At each iteration, among all unfrozen layers, we first select the layer with the largest residual redundancy $R^l$. We then merge the most similar pair of experts within that layer, thereby greedily reducing redundancy. After each merge, we recompute the global entropy. If the updated entropy falls below the threshold $\widehat{E}$, we freeze that layer and prevent further pruning within it. If all layers become frozen before the target budget is reached, we reset the frozen set and continue pruning. This restart mechanism ensures that the pruning process can still reach the target budget while avoiding excessive concentration of pruning in a small number of layers.

\begin{algorithm}[t]
\caption{GRAPE: One-shot entropy-aware global MoE pruning with restart}
\label{alg:entropy_greedy}
\begin{algorithmic}[1]
\Require  Similarity blocks $\{D^{l}\}_{l=1}^{L}$, target experts $K$,
         entropy tolerance $\gamma$
\Ensure   Clusters $\{\mathcal{C}^{l}\}_{l=1}^{L}$ s.t.
          $\sum_{l}\lvert\mathcal{C}^{l}\rvert = K$
\State $\mathcal{C}^{l}\!\gets\!\{\{0\},\dots,\{N{-}1\}\},\;
       R^{l}\!\gets\!\sum_{i\!\neq\! j} D^{l}_{ij}\quad\forall\,l$
\State $E\!\gets\!\operatorname{Entropy}(\{\mathcal{C}^{l}\}),\;
       \widehat{E}\!\gets\!E(1-\gamma),\;
       \mathcal{F}\!\gets\!\varnothing$
\While{$\sum_{l}\lvert\mathcal{C}^{l}\rvert > K$}
    \If{$\mathcal{F} = \{1,\dots,L\}$}  \Comment{all layers frozen}
        \State $\mathcal{F}\gets\varnothing$ \hfill\textit{// restart}
    \EndIf
    \State $l^{\star}\gets\arg\max_{l\notin\mathcal{F}} R^{l}$
    \State $(i^{\star},j^{\star})\gets\arg\max_{i\neq j} D^{l^{\star}}_{ij}$
    \State $\mathcal{C}^{l^{\star}}\!\gets\!
           \operatorname{Union}\bigl(\mathcal{C}^{l^{\star}},i^{\star},j^{\star}\bigr)$
    \State $D^{l^{\star}}_{i^{\star},j^{\star}},\;D^{l^{\star}}_{j^{\star},i^{\star}}\gets 0$
    \State $R^{l^{\star}}\!\gets\!R^{l^{\star}} -
           2D^{l^{\star}}_{i^{\star},j^{\star}}$
    \State $E\!\gets\!\operatorname{Entropy}(\{\mathcal{C}^{l}\})$
    \If{$E<\widehat{E}$}
        \State $\mathcal{F}\gets\mathcal{F}\cup\{l^{\star}\}$ \Comment{freeze}
    \EndIf
\EndWhile
\State \Return $\{\mathcal{C}^{l}\}_{l=1}^{L}$
\end{algorithmic}
\end{algorithm}

\section{Evaluations}

\begin{table*}[htbp]
\centering
\caption{Accuracy (\%) on pruning Mixtral-8x22B, and DeepSeek-MoE with 2 and 4 experts per MoE layer. Each cell reports accuracy in the format $2e/4e$, where we denote $e$ as the number of experts to prune in each layer.}
\label{tab:cut_table}
\resizebox{\textwidth}{!}{
\begin{tabular}{c|c|c|cccccccc}
\toprule
\textbf{Model} & \textbf{Scope} & \textbf{Method} & \multicolumn{4}{c}{\textbf{MMLU}} & \textbf{BoolQ} & \textbf{OpenBookQA} & \textbf{RTE} & \textbf{Average} \\
\cmidrule(lr){4-7}
& & & Humanities & Social Science & STEM & Other & & & & \\
\midrule
\multirow{6}{*}{\textbf{Mixtral-8x22B}} 
& - & {Original~\citep{jiang2024mixtral}} & {68.6} & {84.1} & {67.1} & {78.7} & {87.9} & {35.8} & {71.2} & {70.4} \\
\cmidrule(lr){2-11}
& \multirow{4}{*}{Local}
& Router-guided~\citep{li24merge} & 27.3/22.7 & 25.4/25.8 & 24.4/24.0 & 27.9/23.4 & 62.8/62.7 & 12.8/13.0 & 54.2/49.5 & 33.5/31.6 \\
& & Count-guided~\citep{he2024demystifying} & 58.0/45.7 & 74.9/57.7 & 54.1/42.0 & 70.2/45.7 & 81.5/74.4 & 35.2/27.0 & 69.3/57.4 & 63.3/50.0 \\
& & Enumerate~\citep{lu2024not} & 60.4/53.9 & 78.0/67.2 & 59.5/52.3 & 73.0/64.2 & 87.4/80.5 & 35.0/31.1 & 70.1/67.9 & 66.2/59.6 \\
& & DEK~\citep{zhang2024diversifying} & 62.3/57.8 & 78.5/69.7 & 60.2/51.3 & 73.4/64.2 & 87.6/83.1 & \textbf{35.8}/33.2 & \textbf{71.1}/68.1 & 67.0/61.1 \\
\cmidrule(lr){2-11}
& \multirow{1}{*}{Global} & \textbf{GRAPE (Ours)} & \textbf{64.1}/\textbf{58.4} & \textbf{80.4}/\textbf{72.9} & \textbf{62.7}/\textbf{54.6} & \textbf{75.3}/\textbf{67.9} & \textbf{88.0}/\textbf{84.1}  & 35.2/32.0 & \textbf{71.4}/\textbf{68.5}  & \textbf{68.2}/\textbf{62.6}  \\
\midrule
\multirow{5}{*}{\textbf{Deepseek-MoE}} 
& - & {Original~\citep{dai2024deepseekmoe}} & 40.4 & 47.9 & 36.1 & 49.5  & 77.2 & 32.8 & 66.0 & 50.0 \\
\cmidrule(lr){2-11}
& \multirow{3}{*}{Local}
& Router-guided~\citep{li24merge} & 38.1/34.3 & 46.4/41.6 & 33.9/33.6 & 47.7/43.1 & 71.3/72.1  & 33.2/31.4 & 60.4/60.2 & 47.3/45.2 \\
& & Count-guided~\citep{he2024demystifying} & 38.3/35.9 & 47.4/42.5 & 34.1/32.9 & 47.4/45.6 & 76.2/75.9 & \textbf{33.8}/\textbf{32.4} & 64.9/\textbf{67.5} & 48.9/47.5  \\
& & DEK~\citep{zhang2024diversifying} & 38.7/{39.2} & 47.3/{47.0} & {34.9}/{33.7} & 46.9/{46.6} & 77.4/{76.6} & 32.2/32.2 &64.6/66.4  & 48.8/{48.8} \\
\cmidrule(lr){2-11}
& {Global} & \textbf{GRAPE (Ours)} & \textbf{39.7}/\textbf{39.7} & \textbf{48.3}/\textbf{47.6} & \textbf{35.8}/\textbf{35.3} & \textbf{50.0}/\textbf{50.0}  & \textbf{77.8}/\textbf{77.5}  & {32.0}/31.2 &  \textbf{65.0}/65.5 &  \textbf{49.8}/\textbf{49.5} \\ 
\bottomrule
\end{tabular}}
\end{table*}

\subsection{Experiment setup}
\noindent \textbf{Models.} We study three large-scale MoE models in our experiments: Mixtral-8x22B, Deepseek-MoE-16B, and GPT-oss. Each MoE layer in the Mixtral model contains $8$ experts, with $2$ experts activated per token. Mixtral-8x22B consist of $56$ layers. Deepseek-MoE-16B contains $27$ MoE layers, each comprising $64$ private experts and $2$ shared experts. For each token, $6$ of the $64$ private experts and the $2$ shared experts are activated.GPT-oss (gpt-oss-20b) has 24 MoE layers, where 4 of 32 experts in each layer are activated for each token.

\begin{table*}[htbp]
\centering
\caption{Accuracy (\%) on  pruning GPT-OSS with 2 and 4 experts per MoE layer. Each cell reports accuracy in the format $2e/4e$, where we denote $e$ as the number of experts to prune in each layer.}
\label{tab:cut_table_gptoss}
\resizebox{\textwidth}{!}{
\begin{tabular}{c|c|c|ccccc}
\toprule
\textbf{Model} & \textbf{Scope} & \textbf{Method} &
\textbf{MMLU} & \textbf{BoolQ} & \textbf{OpenBookQA} & \textbf{RTE} & \textbf{Average} \\
\midrule
\multirow{6}{*}{\textbf{GPT-oss}}
& -- & \textcolor{gray}{Original~\citep{dai2024deepseekmoe}}
& 85.6 & 88.7 & 93.0 & 92.8 & 90.0\\
\cmidrule(lr){2-8}
& \multirow{4}{*}{Local}
& Router-guided~\citep{li24merge}
& 83.4/81.3 & 88.9/87.3 & 89.6/85.6 & 90.6/89.5 & 88.1 / 85.9 \\
&  & Count-guided~\citep{he2024demystifying}
& 84.8/82.5 & 88.2/89.3 & 93.6/93.2 & 92.4/91.3 &  89.8 / 89.1\\
&  & Enumerate~\citep{lu2024not}
& 83.7/82.5 & 88.8/88.7 & 93.4/92.8 & 93.5/92.1 &   89.9 / 89.0\\
&  & DEK~\citep{zhang2024diversifying}
& 83.6/80.8 & 89.0/88.1 & 94.6/92.4 & 92.4/90.6 & 89.9 / 88.0 \\
\cmidrule(lr){2-8}
& Global & \textbf{GRAPE (Ours)}
& 85.3/83.4 & 89.0/89.0 & 94.2/93.6 & 92.6/91.9 & 90.3 / 89.5 \\
\bottomrule
\end{tabular}}
\end{table*}

\noindent \textbf{Baselines and implementations:} We compare GRAPE  with four recent MoE pruning approaches~\citep{he2024demystifying, li24merge, lu2024not, zhang2024diversifying}. For locally based pruning baselines, we include the \textit{router-guided} method~\citep{li24merge}, which identifies similar experts using router information; \textit{Expert Trimming}~\citep{lu2024not}, a frequency-based approach referred to as the \textit{count-guided} strategy; and a loss-based pruning method from the same work, denoted as \textit{Enumerate}. We also consider \textit{DEK} (Diversifying Expert Knowledge)~\citep{zhang2024diversifying}, which detects redundant experts based on their output representations.

For Mixtral and DeepSeek-MoE, the models are prompted to directly generate the final answer. In contrast, GPT-OSS follows its default reasoning style, where the model first produces intermediate reasoning steps before generating the final answer. We set the reasoning effort of GPT-OSS to \textit{medium} and randomly sample $1000$ examples from MMLU for evaluation due to computational cost. For Mixtral and DeepSeek-MoE, we evaluate on the full MMLU dataset. For all other datasets, we use the complete evaluation sets.

All baseline methods perform uniform pruning by removing the same number of experts from each MoE layer. In contrast, our method maintains the same overall pruning budget but adaptively determines the number of experts to prune in each layer according to layer-wise redundancy. To ensure a fair comparison in the task-agnostic setting, we disable all fine-tuning stages for all methods.

\subsection{Experiment Results}

\cref{tab:cut_table,tab:cut_table_gptoss} report the pruning accuracy of Mixtral-8x22B, DeepSeek-MoE, and GPT-OSS. Overall, GRAPE consistently achieves the best average accuracy under both pruning settings. On Mixtral-8x22B, GRAPE reaches average accuracies of 68.2 (2e) and 62.6 (4e), outperforming the strongest local baseline by 1.79\% and 2.45\%, respectively. On DeepSeek-MoE, the gains are smaller but consistent, with GRAPE achieving 49.8 (2e) and 49.5 (4e), corresponding to relative improvements of 1.84\% and 1.43\% over the strongest local baseline. On GPT-OSS, GRAPE obtains 90.3 (2e) and 89.5 (4e), improving over the strongest local baseline by 0.44\% and 0.45\%, respectively. These results show that allocating pruning budgets globally according to cross-layer redundancy leads to consistently better accuracy-compression trade-offs than uniform layer-wise pruning. More experiment results on Mixtral-8x7B and Qwen-MoE are provided in \cref{sec:appendix_more_results}.

\section{Conclusion}

We propose a global pruning strategy for sparse Mixture-of-Experts  models that dynamically allocates pruning budgets based on cross-layer redundancy, enabling more efficient expert removal. Our approach consistently outperforms strong baselines, demonstrating its effectiveness in preserving performance under constrained memory budgets. However, experiments on Deepseek-MoE reveal that severe imbalance in layerwise redundancy can cause global pruning to collapse the model. These results highlight both the promise and the limitations of globally guided pruning, calling for future work on more adaptive and robust strategies, including the design of a suitable metric to evaluate the MoE layer redundancy,  for compressing MoEs from a global perspective.


\bibliography{custom}

\newpage
\begin{appendices}

\begin{figure*}
    \centering
    \includegraphics[width=\linewidth]{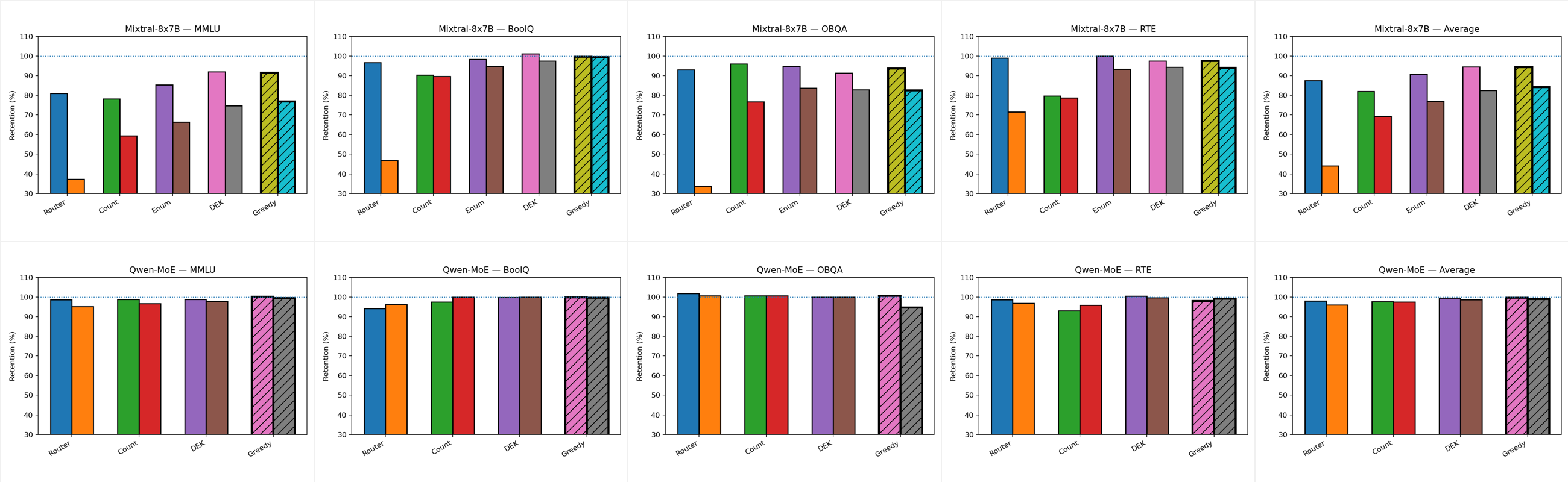}
    \caption{Results of Mixtral-8x7B and Qwen-MoE.}
    \label{fig:retention_appendix}
\end{figure*}
\section{More Results on MoE Pruning}
\label{sec:appendix_more_results}

This section provides additional experimental results on \textbf{Mixtral-8$\times$7B} and \textbf{Qwen-MoE}, which are omitted from the main text.
We follow the same experimental protocol and evaluation metric as in the main experiments and report results using \textbf{retained performance}, defined as the ratio between the accuracy of the pruned model and that of the original model.

\cref{fig:retention_appendix} presents task-wise retained performance for Mixtral-8$\times$7B and Qwen-MoE. Consistent with the main results, Global Greedy achieves the highest or near-highest retained performance across tasks for both models.
On Mixtral-8$\times$7B, the advantage of Global Greedy becomes particularly clear under four-expert pruning, where uniform layer-wise pruning methods suffer substantial retention drops, especially on MMLU.
On Qwen-MoE, where pruning is generally less destructive, Global Greedy still provides consistently strong retention across all tasks and maintains the best overall average performance.

These results further confirm that allocating pruning budgets globally based on cross-layer redundancy leads to more stable performance preservation than uniform per-layer pruning, even for smaller or less redundant MoE models.

\end{appendices}

\end{document}